\documentclass[runningheads]{llncs}
\usepackage{graphicx}
\usepackage[utf8]{inputenc}
\usepackage[english]{babel}

\usepackage{algorithm}

\usepackage{algpseudocode}
\usepackage{amsmath}
\usepackage{setspace}

\usepackage{hyperref}
\begin{document}
\title{MetalGAN: a Cluster-based Adaptive Training for Few-Shot Adversarial Colorization}
\titlerunning{Cluster-based Adaptive Training for Few-Shot Adversarial Colorization}
%
\author{Tomaso Fontanini\orcidID{0000-0001-6595-4874} \and
Eleonora Iotti\orcidID{0000-0001-7670-2226} \and
Andrea Prati\orcidID{0000-0002-1211-529X}}
\authorrunning{T. Fontanini, et al.}
%
\institute{IMP Lab, Department of Engineering and Architecture, University of Parma
\email{tomaso.fontanini@studenti.unipr.it}\\
\email{\{eleonora.iotti, andrea.prati\}@unipr.it}}
\maketitle              

\begin{figure}
    \centering
    \includegraphics[width=\textwidth]{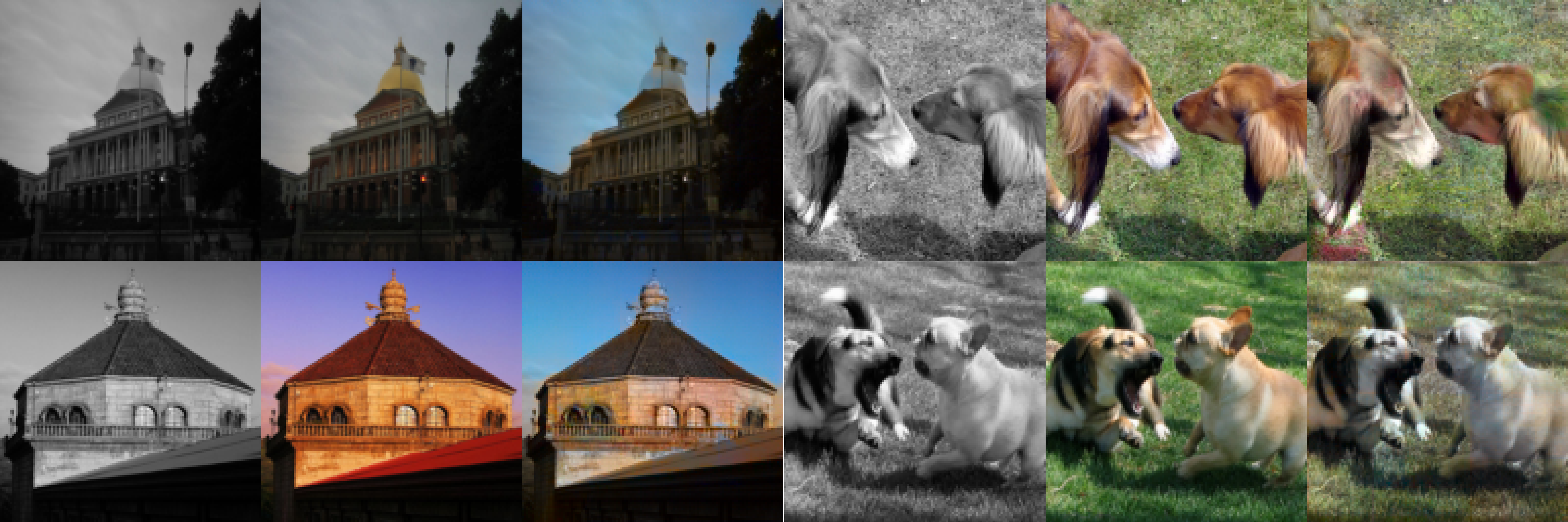}
    \caption{Example images generated using MetalGAN for 100-epochs, and 100-meta-iterations. From left to right: gray scale image, ground truth, output of the network. The example images belong to two different clusters.}
    \label{fig:meta_net}
\end{figure}

\begin{abstract}
In recent years, the majority of works on deep-learning-based image colorization have focused on how to make a good use of the enormous datasets currently available. What about when the data at disposal are scarce? The main objective of this work is to prove that a network can be trained and can provide excellent colorization results even without a large quantity of data. The adopted approach is a mixed one, which uses an adversarial method for the actual colorization, and a meta-learning technique to enhance the generator model. Also, a clusterization \emph{a\nobreakdash-priori} of the training dataset  ensures a task-oriented division useful for meta-learning, and at the same time reduces the per-step number of images. This paper describes in detail the method and its main motivations, and a discussion of results and future developments is provided.

\end{abstract}
\vspace{-0.3cm}
\section{Introduction}
\vspace{-0.2cm}

The \emph{automatic image colorization} task is an image processing problem that is fundamental and extensively studied in the field of computer vision.
The task consists in creating an algorithm that takes as input a gray-scale image and outputs a colorized version of the same image. The challenging part is to colorize it in a plausible and well-looking way.
Many systems were developed over the years, exploiting a wide variety of image processing techniques, but recently, the image colorization problem, as many other problems in computer vision, was approached with deep-learning methods.
Colorization is a \emph{generative} problem from a machine learning perspective.
Generative techniques, such as \emph{Generative Adversarial Networks} (\emph{GANs}) \cite{goodfellow2014generative}, are then suitable to approach such a task. 
In particular, \emph{conditional GANs} (\emph{cGANs}) models seem especially appropriate to this purpose, since their structure allows the network to learn a mapping from an image $x$ and (only if needed) a random noise vector $z$ to an output generated image $y$. 
On the contrary, standard GANs only learn the mapping from the noise $z$ to $y$.

As many deep-learning techniques, the training of a GAN or a cGAN needs a large amount of images. Large datasets usually grant a great diversity among images, allowing the network to better generalize its results.
Nevertheless, having a huge number of images is often not feasible in real-world applications, or simply it requires too much storage space for an average system, and high training computational times. 
Hence, porting the current deep-learning colorization technologies to a more accessible level and achieving a better understanding of the colorization training  process are eased by using a smaller dataset.

For these reasons, one of the aims of this work is to achieve good performances in the colorization task using a little number of images compared to standard datasets.
In \textit{few-shot learning}, a branch of the deep-learning field, the goal is to learn from a small number of inputs, or from one single input in the ideal case (\textit{one-shot learning}): the network is subject to a low quantity of examples, and it has to be capable to infer something when posed face-to-face to a new example. 
This problem underpins a high generalization capability of the network, which is a very difficult task and an open challenging problem in deep networks research.

Recently, some novel interesting ideas highlight a possible path to reach a better generalization ability of the network. These ideas are based on the concept of learning to learn, i.e., adding a meta-layer of learning information above the usual learning process of the network. 
The generalization is achieved by introducing the concept of \emph{tasks distribution} instead of a single task, and the concept of \emph{episodes} instead of instances. A tasks' distribution is the family of those different tasks on which the model has to be adapted to. Each task in the distribution has its own training and test sets, and its own loss function.
A meta-training set is composed of training and test images samples, called episodes, belonging to different tasks.
During training, these episodes are used to update the initial parameters (weights and bias) of the network, in the direction of the sampled task. 
Results of meta-learning methods investigated in literature are encouraging and obtain good performances on some few-shot datasets.
For this reason and since the goal of this work is to colorize images with a few number of examples, a meta-learning algorithm to tune the network parameters on many different tasks was employed.
The chosen algorithm is Reptile \cite{nichol2018first}, and it was combined with an adversarial colorization network composed by a Generator $G$ and a Discriminator $D$. 
In other words, the proposed method approaches the colorization problem as a meta-learning learning one.
Intuitively, Reptile works by randomly selecting tasks, then it trains a fast network on each task, and finally it updates the weights of a slow network.

In this proposal, tasks are defined as clusters of the initial dataset. In fact, a typical initial dataset is an unlabeled dataset that contains a wide variety of images, usually photographs. In this setting, for example, a task could be to color all seaside landscape, and another could be to color all cats photos. Those tasks refer to the same problem and use the same dataset, but they are very different at a practical level. A very large amount of images could overwhelm the problem, showing as much seasides and cats as the network needs in order to differentiate between them. The troubles start when only a small dataset is available. As a matter of fact, such a dataset could not have the suitable number of images for making the network learning how to perform both the two example colorizations decently. The idea is to treat different classes of images as different tasks.
For dividing tasks, features were extracted from the dataset using a standard approach---e.g., a Convolutional Neural Network (CNN)---and the images were clusterized through K-means. Each cluster is thus considered as a single task.
During training, Reptile tunes the network $G$ on the specific task corresponding to an input query image and therefore it adapts the network to a specific colorization class.

The problems and main questions that emerge in approaching a few-shot colorization are various. First of all, how the clusterization should be made in order to generate a coherent and meaningful distribution of tasks? Does a task specialization really improve the colorization or the act of automatically coloring a photo is independent from the subject of the photo itself?
Second, how the meta-learning algorithm should be combined with cGAN training, also to prevent overfitting the generator on few images? 
And last, since the purpose of the work is not to propose a solution to the colorization problem in general, but to propose a method that substantially reduce the amount of images involved in training without---or with minor---losses in state-of-the-art results, how to evaluate the actual performance of the network compared to other approaches?
In particular, what are the factors that should be taken in account to state an enhancement, not in the proper colorization, but in few-shot colorization?
In the light of these considerations, the contributions of this work are summarized as follows:
\begin{itemize}
    \item A new architecture that combines meta-learning techniques and cGAN called \textit{MetalGAN} is proposed, specifying in detail how the generator and the discriminator parameters are updated;
    \item A clusterization and a novel algorithm are described and their ability to tackle image-to-image translation problems is highlighted;
    \item An empirical demonstration that a very good colorization can be achieved even with a small dataset at disposal during training is provided by showing visual results;
    \item A precise comparison between two modalities (i.e. our algorithm and only cGAN training) is performed at experimental time, using the same network model and hyper-parameters.
\end{itemize}


\vspace{-0.3cm}
\section{Related Work}
\label{sec:related}
\vspace{-0.2cm}

\subsubsection{Image retrieval:}
Since we need the clusterization to be as accurate as possible we reserved a particular attention to the recent image retrieval techniques that focus on obtaining optimal descriptors. Recently, deep learning allowed to greatly improve the feature extraction phase of image retrieval. Some of the most interesting papers on the subject are \cite{razavian2016visual,gong2014multi,babenko2014neural,yue2015exploiting,reddy2015object} and, in particular, MAC descriptors \cite{tolias2015particular}, that we ended up using.

\subsubsection{Conditional GANs:} 
When a GAN generator is not only conditioned with a random noise vector, but also with more complex information like text \cite{reed2016generative}, labels \cite{mirza2014conditional}, and especially images, the model to use is a \textit{conditional} GANs (\textit{cGANs}). cGANs allow a better control over the output of the network and thus are very suitable in a lot of image generation tasks. In particular, cGANs conditioned on images were used both in a paired \cite{isola2017image} and unpaired \cite{zhu2017unpaired} way, to produce complex texture \cite{xian2018texturegan}, to colorize sketches \cite{sangkloy2017scribbler} or images \cite{cao2017unsupervised} and more recently to produce outstanding image synthesis results \cite{wang2018high,park2019semantic}.
In this work, the output must be conditioned by the input gray-scale image, in order to train the network at only generating the colors of the image but not shapes, or the image itself. 

\subsubsection{Meta-learning:}
The most relevant meta-learning studies for this work are the Model-Agnostic Meta-Learning (MAML) \cite{finn2017model} algorithm and Reptile \cite{nichol2018first} ones. In particular, we incorporate the Reptile algorithm inside the training phase, allowing the parameters of the generator to be updated in the same fashion as Reptile works. 
A similar work using MAML is MetaGAN \cite{zhang2018metagan}, where a generator is used to enhance classification models in order to discriminate between real and fake data, providing generated samples for a task. The main purpose of MetaGAN is not to improve a generative network, but to perform a better few-shot classification, using generated images to sharpen the decision boundary of the problem. On the contrary, in our approach, the generator is fed with task-related images, and the meta-learner is used to enhance the generator itself, instead of a few-shot classifier. 
Both MAML and Reptile are based on hyper-parameterized gradient descent, and they learn how to initialize network parameters. Other types of meta-learners work differently. For example, there are many algorithms that learn how to parameterize the optimizer of the network \cite{hochreiter2001learning,ravi2016optimization}, or in other cases the optimizer itself is a network \cite{li2017learning,andrychowicz2016learning,wichrowska2017learned}. Moreover, one of the most general approach is to use a recurrent neural network trained on the episodes of a set of tasks \cite{santoro2016meta,mishra2017simple,duan2016rl,wang2016learning}. The most interesting result of these meta-learners is the achievement of high performance on small datasets \cite{Rezende:2016:OGD:3045390.3045551,vinyals2016matching,kiran2018zero}, or datasets used for few-shot learning (e.g., Omniglot) \cite{lake2015human}.

\vspace{-0.3cm}
\section{Algorithm}
\label{sec:algorithm}
\vspace{-0.2cm}

This section goes in detail within the algorithm we propose. Therefore, each subsection focuses on a different aspect of the method. Then, the complete architecture is explained.

\vspace{-0.3cm}
\subsection{Clusterization of the dataset}
\vspace{-0.2cm}

In order to exploit Reptile for image colorization we need to treat our image dataset as it would be composed by a series of separate tasks. For this reason we extract features from each image in the dataset using \textit{activation\_43} layer of Resnet50. Then, we calculate MAC descriptors by applying max pooling and L2 normalization on the features. Having these MAC descriptors set $F$, we first apply Principal Componet Analysis (PCA) to reduce features dimension from 2048 to 512 and then apply K-means. K-means produces $k$ clusters, and therefore it divides the dataset in $k$ tasks. 

\begin{figure}[t]
    \centering
    \includegraphics[width=0.5\linewidth]{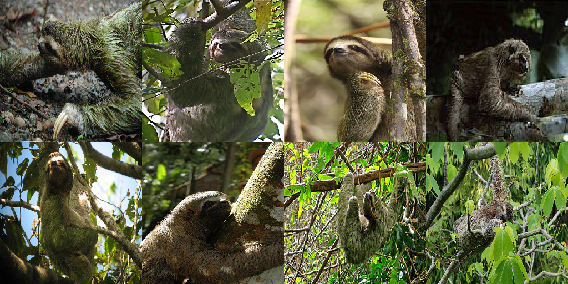}
    \caption{Some of the results of the clusterization. It is evident how all the images have lots of features in common.}
    \label{fig:clusters}
\end{figure}
Hence, we expect to find, in each of these clusters, images which are similar to each other, accordingly to their features. For example, a cluster could contain images with grass, another one images with pets and so on and so forth. A visual proof of this assumption is showed in Fig. \ref{fig:clusters}.

\vspace{-0.3cm}
\subsection{cGAN}
\vspace{-0.2cm}

As generator architecture, we choose the U-net \cite{ronneberger2015u} which is one of the most common for this type of task and we built the discriminator following the classic DCGAN architecture \cite{radford2015unsupervised}, i.e., having each modules composed by Convolutions, Batch Normalization and ReLU layers.
Lab is the color space used in this work,  because is the one that best approximate human vision and therefore the generator takes as input a grayscale image $x_i$ (the $L$ channel) and outputs the $ab$ channels. Then, we concatenate input and outputs and obtain the final results.

We use L1 loss to model the low-frequencies of our output images and adversarial loss to model the high-frequencies in a similar way of the pix2pix architecture proposed by Isola \textit{et al.} \cite{isola2017image}. 

Therefore, our objective function became:

\begin{equation}\label{eqn:loss}
\mathcal{L} = \textbf{w}_\textbf{adv}\mathcal{L}_\textbf{adv} + \textbf{w}_\textbf{L1}\mathcal{L}_\textbf{L1}
\end{equation}
where $\textbf{w}_\textbf{adv}$ and $\textbf{w}_\textbf{L1}$ are weights assigned to the different losses, because we want L1 loss to be more effective than adversarial loss during training.

\vspace{-0.3cm}
\subsection{Meta-learning}
\vspace{-0.2cm}

As previously briefly mentioned, we approached the generator training with a Reptile meta-learner. 
This means that, once a task had been chosen, for a fixed number of meta-iterations, the task is sampled and the gradient of the generator loss function \eqref{eqn:loss} is evaluated to perform a SGD step of optimization.
Fixed the initial generator parameters as $\theta_G$, the inner-loop training defines a sequence $\left(\tilde{\theta}_G^{(j)}\right)_{j = 0}^{N_{\mathrm{meta-iter}}}$, where $\tilde{\theta}_G^{(0)} = \theta_G$.
Hence it updates the $\tilde{\theta}_G^{(j)}$ parameters in the direction of the task.
Once the inner-loop is completed, the parameter are re-aligned with the Reptile rule:
\begin{equation}
    \theta_G \gets \theta_G + \lambda_{ML}\left(\tilde{\theta}^{(N_{\mathrm{meta-iter}})}_G - \theta_G\right)
\end{equation}
where $\lambda_{ML}$ is the stepsize hyperparameter of Reptile.

\vspace{-0.3cm}
\subsection{Complete architecture of the system}
\vspace{-0.2cm}

The \textit{MetalGAN} training process is detailed in Algorithm \ref{metal_alg}.
\begin{algorithm}[h]
\small
\setstretch{1}
\caption{MetalGAN algorithm}\label{alg:metalgan}
\begin{algorithmic}[1]
\For{$epoch \; \mathbf{in} \; 0 \dots N_{\mathrm{epochs}}$}
    \For{$q_i \; \textbf{in} \; Q$}
    
        \State $K(q_i) \gets$ retrieve\_clusters($q_i$)
        \State $\tau(q_i) \gets$ get\_task\_from\_cluster($K(q_i)$)
        
        \For{$j \; \textbf{in} \; 0 \dots N_{\mathrm{meta-iter}}$}
        
            \State sample $\langle \mathrm{input,target} \rangle$ from task $\tau(q_i)$
            \State $\varepsilon_{\mathrm{GAN}} \gets \nabla_{\theta_D}\mathcal{L}_\textbf{adv}$(D(G(input)), label\_real)
            \State $\varepsilon_{\mathrm{L1}} \gets \nabla_{\theta_G}\mathcal{L}_\textbf{L1}$(D(G(input)), target)
            \State $\varepsilon_G \gets \textbf{w}_\textbf{adv}\varepsilon_{\mathrm{GAN}} + \textbf{w}_\textbf{L1}\varepsilon_{\mathrm{L1}}$
            \Comment{calculates loss gradient}
            \State $\tilde{\theta}^{(j)}_G \gets \tilde{\theta}^{(j-1)}_G -\lambda_G\varepsilon_G $ \Comment{updates inner-loop generator parameters}
        
        \EndFor
        
        \State $\theta_G \gets \theta_G + \lambda_{ML}\left(\tilde{\theta}^{(N_{\mathrm{meta-iter}})}_G - \theta_G\right)$
        \Comment{updates generator parameters}
        
        \ForAll{$\langle \mathrm{input,target} \rangle \; \mathbf{in} \; \tau(q_i)$} 
        
            \State $\varepsilon_{D_{\mathrm{real}}} \gets \nabla_{\theta_D}\mathcal{L}_\textbf{adv}$(D(target), label\_real)
            \State $\varepsilon_{D_{\mathrm{fake}}} \gets  \nabla_{\theta_D}\mathcal{L}_\textbf{L1}$(D(G(input)), label\_fake)
            \State $\varepsilon_D \gets \varepsilon_{D_{\mathrm{real}}} + \varepsilon_{D_{\mathrm{fake}}} $
            \Comment{calculates discriminator loss gradients}
            \State $\theta_D \gets \theta_D - \lambda_D \varepsilon_D$
            \Comment{update discriminator parameters}
        
        \EndFor

    \EndFor
\EndFor

\end{algorithmic}
\label{metal_alg}
\end{algorithm}
The algorithm is parameterized by the number of epochs $N_{\mathrm{epochs}}$, the number of meta-iterations $N_{\mathrm{meta-iter}}$, the generator and discriminator learning rates $\lambda_G$ and $\lambda_D$, the Reptile stepsize parameter $\lambda_{ML}$, and the loss weights $\textbf{w}_\textbf{adv}$ and $\textbf{w}_\textbf{L1}$.
During training, we randomly select a query set $Q = \{q_0,\dots,q_z\}$. Each query $q_i$ corresponds to a single cluster $K(q_i)$. It is worth noting that two queries could point to the same cluster. 
Having this set, we are able to pick $z$ different images at each epoch by sampling the task $\tau(q_i)$ and to update the generator $G$ as showed in Fig. \ref{fig:meta_net}.
\begin{figure}[H]
    \centering
    \includegraphics[width=0.8\linewidth]{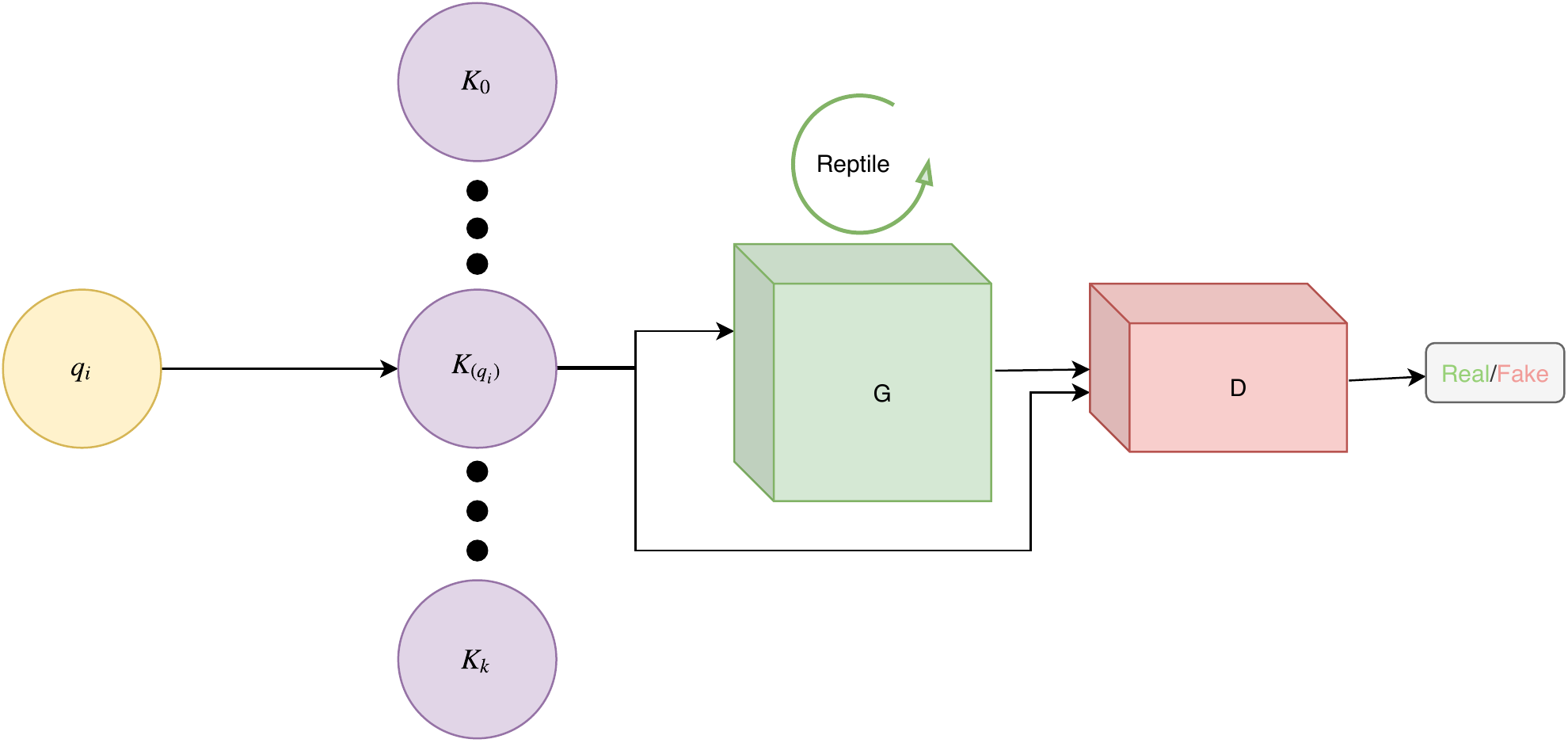}
    \caption{The MetalGAN architecture: the query $q_i$ points to a cluster $K(q_i)$ that is used as a task to train the Generator $G$ with Reptile.}
    \label{fig:meta_net}
\end{figure}
The generator is updated by evaluating gradients of its loss functions (adversarial loss $\mathcal{L}_{\mathrm{adv}}$ and L1 loss $\mathcal{L}_{\mathrm{L1}}$), and by adding them to obtain the error $\varepsilon_G$.
Then, the network parameters obtained in the inner-loop $\tilde{\theta}_G^{N_{\mathrm{meta-iter}}}$ are used to update the outer-loop generator parameters $\theta_G$.
In the last step, all images of the task $\tau(q_i)$ are used to train the discriminator, calculating the gradients of the discriminator adversarial and L1 losses, and adding them to obtain the discriminator error $\varepsilon_D$. The discriminator parameters $\theta_D$ are updated consequently.

\vspace{-0.3cm}
\section{Experimental Results}
\label{sec:results}
\vspace{-0.2cm}

For our experiments we choose a slightly modified version of Mini-Imagenet \cite{ravi2016optimization}. Since our goal is not classification, we create our training and test set using only images from the 64 classes contained in the training section of Mini-Imagenet. 
The total number of images in the dataset is $38392$. 
We define two sets of experiments: the first one consists in training the cGAN without the use of Reptile and the second one introduces Reptile and the features clusterization. For both of them we set $\textbf{w}_\textbf{adv} = 1$ and $\textbf{w}_\textbf{L1} = 10^2$. Learning rates of both the generator and the discriminator were set to $\lambda_G = \lambda_D = 10^{-4}$.
For K-means clusterization, the parameter $k$ was set to 64 in order to have clusters as much disjoint as possible. 
For Reptile, we use 100 \textit{meta-iter}, and a stepsize $\lambda_{ML} = 10^{-3}$. The 10\% of the dataset images are used as query images. The number of epochs was set to 200.
All tests have been executed on a GPU Nvidia 1080 Ti.

\vspace{-0.3cm}
\subsection{cGAN results}
\vspace{-0.2cm}

\begin{figure}[t]
    \centering
    \includegraphics[width=\linewidth]{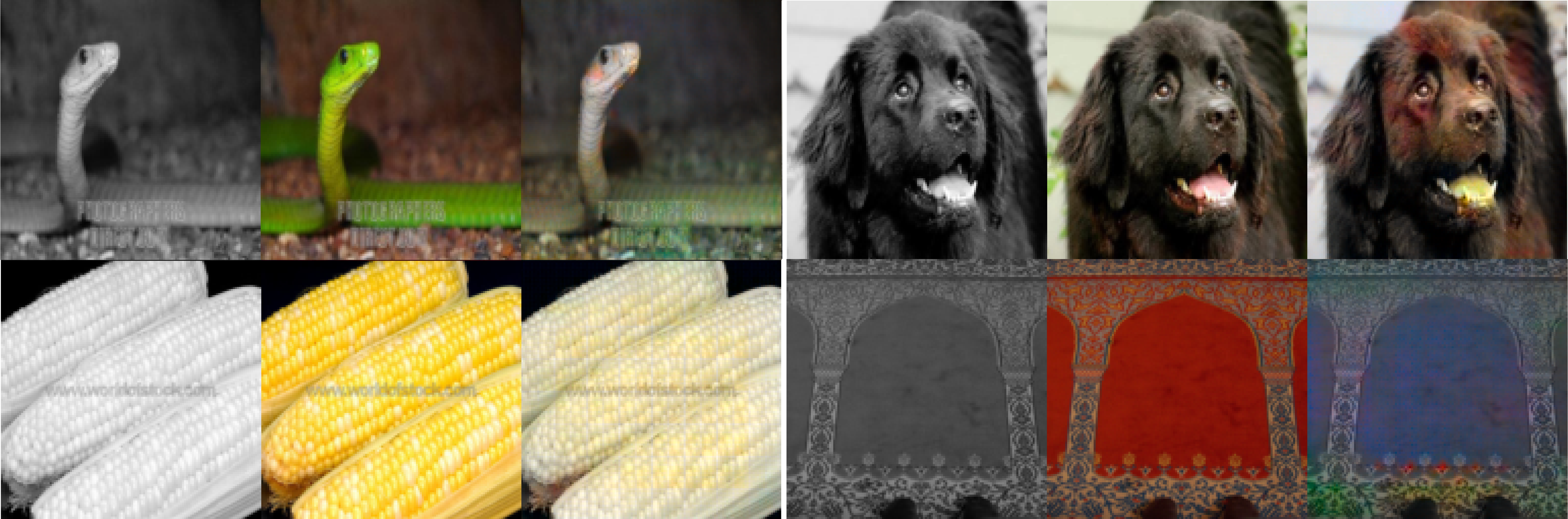}
    \caption{Results obtained using the cGAN only. Each group of three images is composed of the input of the network (grayscale image), the ground truth, and the output of the network.}
    \label{fig:soloGAN_res}
\end{figure}
In Fig. \ref{fig:soloGAN_res} are reported some results produced after the training of the cGAN without the clusterization and without Reptile, i.e., with a standard adversarial algorithm. The training data at disposal are very scarce ($\sim$38k images compared to 1.3M of the whole Imagenet dataset) and, for this reason, the network is not able to produce compelling results. 
In particular, the network often fails to understand the difference between foreground and background objects and therefore it applies the colors without following edges and borders. 
In general, for the cGAN is very difficult to propagate the color correctly and is more common the tendency to apply uneven patches of color. 
Finally, due to the scarcity of data, the network cannot generalize in an acceptable way and hence the colors in the outputs are not sharp, but, on the contrary, the produced results are very blurry and often colors are applied almost randomly.

\vspace{-0.3cm}
\subsection{MetalGAN results}
\vspace{-0.2cm}

\begin{figure}[t]
    \centering
    \includegraphics[width=\linewidth]{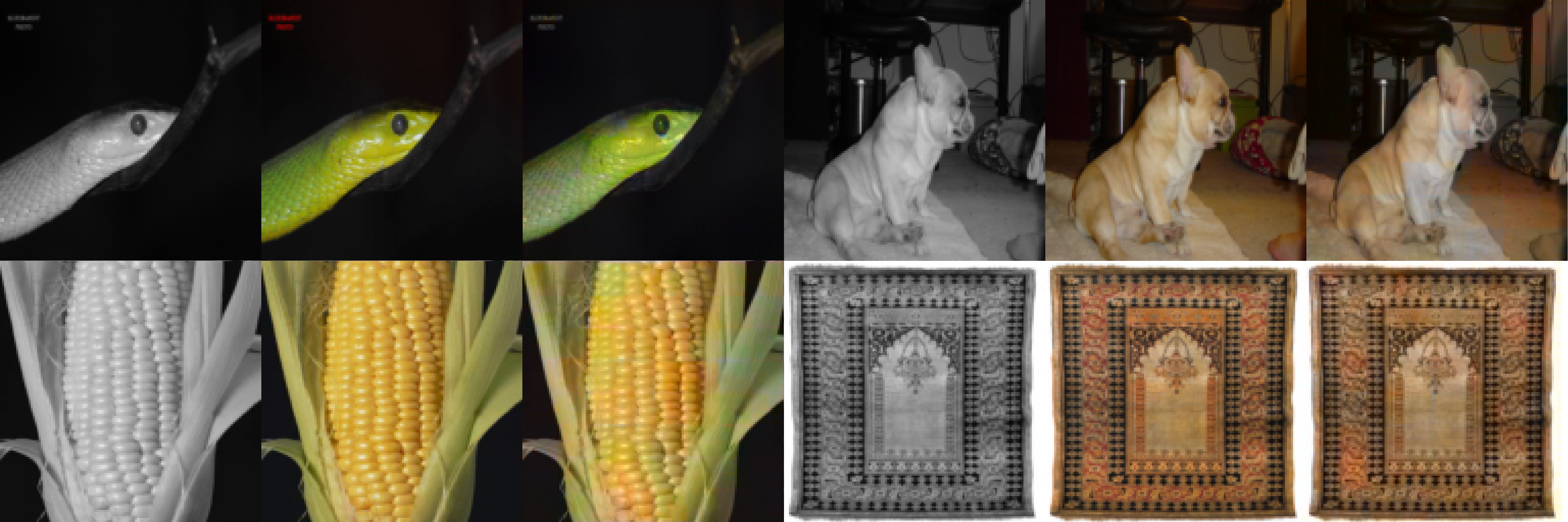}
    \caption{Results of MetalGAN. Each of three images consists of the grayscale input given to the network, the ground truth, and the output of the network. The four represented images belong to different clusters.}
    \label{fig:metalGAN_res}
\end{figure}

Results of MetalGAN are showed in Fig. \ref{fig:metalGAN_res}. It is immediately evident how Reptile improves the results of the cGAN. In particular, colors are sharper and more bright. The reason is that Reptile tunes the generator on each cluster and therefore allows the network to focus more on the more predominant colors present in each task and, as a consequence, even with few examples the produced results are compelling and plausible. For example, in a task with lots of images containing grass or plants there will be an abundance of different shades of green and thus the network will learn very quickly to reproduce similar colors over the test set.
On the contrary, an image that is very different from the majority of images in the rest of its task could be colorized poorly. This problem, however, is not very frequent since the difference has to be very large in order to produce nasty results.

Other examples can be found at \url{implab.ce.unipr.it/?page_id=1011}.

\vspace{-0.3cm}
\subsection{Quantitative evaluation}
\vspace{-0.2cm}

In order to evaluate the quality of the generated samples, we used the Inception Score \cite{inceptionscore2016}, because it is a very good metric to simulate human judgement.
We calculated the Inception Score of generated images using both cGAN and MetalGAN (see Table~\ref{tab:res}). The score also measures the diversity of the generated images, so a high score is better than a lower one. The MetalGAN approach significantly improves standard cGAN score.

\begin{table}[t]
    \centering
    \begin{tabular}{|c|c|c|}
        \hline
         \bf Dataset & \bf Mean & \bf Std \\
         \hline
         cGAN & 3.20 & 0.83 \\
         \hline
         MetalGAN & 9.16 & 1.12 \\
         \hline
    \end{tabular}
    \caption{The Inception Scores are computed on generated images from the MiniImageNet dataset, mean and standard deviation are reported for both cGAN and MetalGAN results.}
    \label{tab:res}
\end{table}

\vspace{-0.3cm}
\section{Conclusions}
\vspace{-0.2cm}

In normal adversarial generative settings, having few images at disposal during training produces a complete failure in the colorization.
In this paper, we proposed a novel architecture which mix adversarial training with meta-learning techniques, called MetalGAN.
As shown by experimental results, even with few images the network trained with MetalGAN was able to produce a well-looking colorization. The clusterization of the dataset and the use of clusters as tasks help at directing the colorization to the most probable suitable colors for the image, and meta-learning allows to train the network on few examples. As future developments, we plan to include the discriminator in the meta-learning training phase, and to test the method on other small datasets in order to prove the generalization capability of the proposed MetalGAN architecture.

%
%
%
\bibliographystyle{splncs04}
\bibliography{mybibliography}

\end{document}